\documentclass[runningheads]{llncs}

\usepackage{comment}
\usepackage{color}

\usepackage{amsmath}
\usepackage{amssymb}
\usepackage{helvet} % DO NOT CHANGE THIS
\usepackage{courier}  % DO NOT CHANGE THIS
\usepackage[hyphens]{url}  % DO NOT CHANGE THIS
\usepackage{graphicx} % DO NOT CHANGE THIS
\usepackage[utf8]{inputenc} % allow utf-8 input
\usepackage[T1]{fontenc}    % use 8-bit T1 fonts
\usepackage{url}            % simple URL typesetting
\usepackage{booktabs}       % professional-quality tables
\usepackage{amsfonts}       % blackboard math symbols
\usepackage{nicefrac}       % compact symbols for 1/2, etc.
\usepackage{microtype}      % microtypography
\usepackage{subfig}
\usepackage{textcomp}
\usepackage{caption,sidecap}
\usepackage[table]{xcolor}
\usepackage{nicefrac}
\usepackage{color}
\definecolor{Gray}{gray}{0.85}
\newcommand{\graphicalmodel}{
\begin{SCfigure}[1][!ht]
\centering
 \vspace{-.2in}
    \includegraphics[width=0.5\textwidth, trim=3.1in 0.8in 2.9in 0.8in, clip=true]{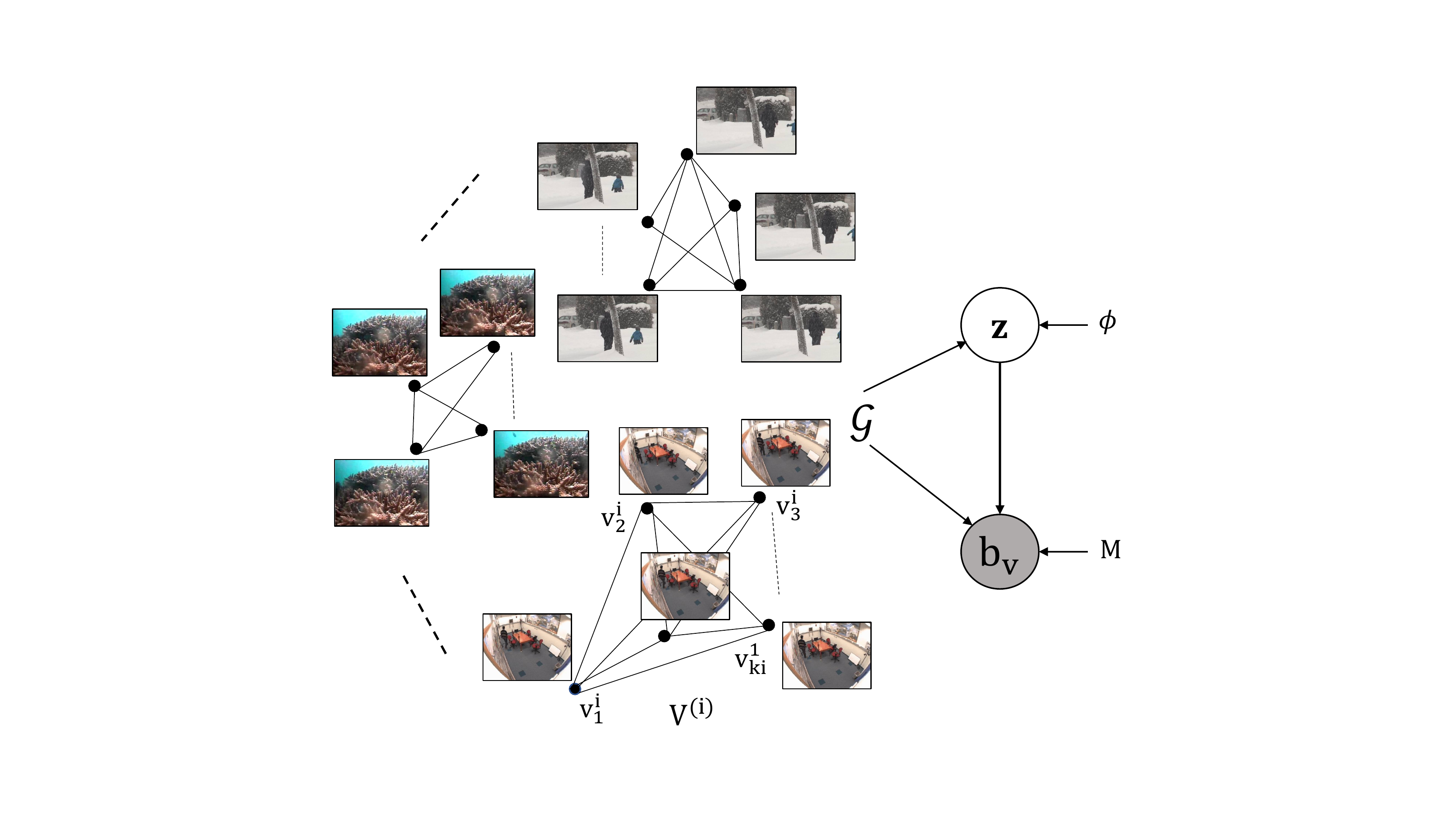}
    \hspace{-.2in}
    \caption{\footnotesize Graphical representation  of the generative process in G-LBM. Given the video dataset $\{\mathbf{v}_1, \dots , \mathbf{v}_n\}$, we construct a neighbourhood graph $\mathcal{G}$ in which video frames from the same scene create a clique in the graph, as $V^{(i)} = [\mathbf{v}^i_1, \dots , \mathbf{v}^i_{ki}]$. The distribution over latent process $\mathbf{z}$ is controlled by the graph $\mathcal{G}$ as well as the parameters of the non-linear mapping $\phi$. The latent process $\mathbf{z}$ along with the motion mask $M$ determine the likelihood of the background $\mathbf{b_v}$ in video frames $\mathbf{v} \in \{\mathbf{v}_1, \dots , \mathbf{v}_n\}$.}
    \label{fig:grmodel}
    \vspace{-.1in}
\end{SCfigure}
}

\newcommand{\networkarchitecture}{
\begin{figure*}[!ht]
\centering
 \vspace{-.1in}
    \includegraphics[width=0.6\textwidth, trim=0.2in 0.3in 0.2in 0.3in, clip=true]{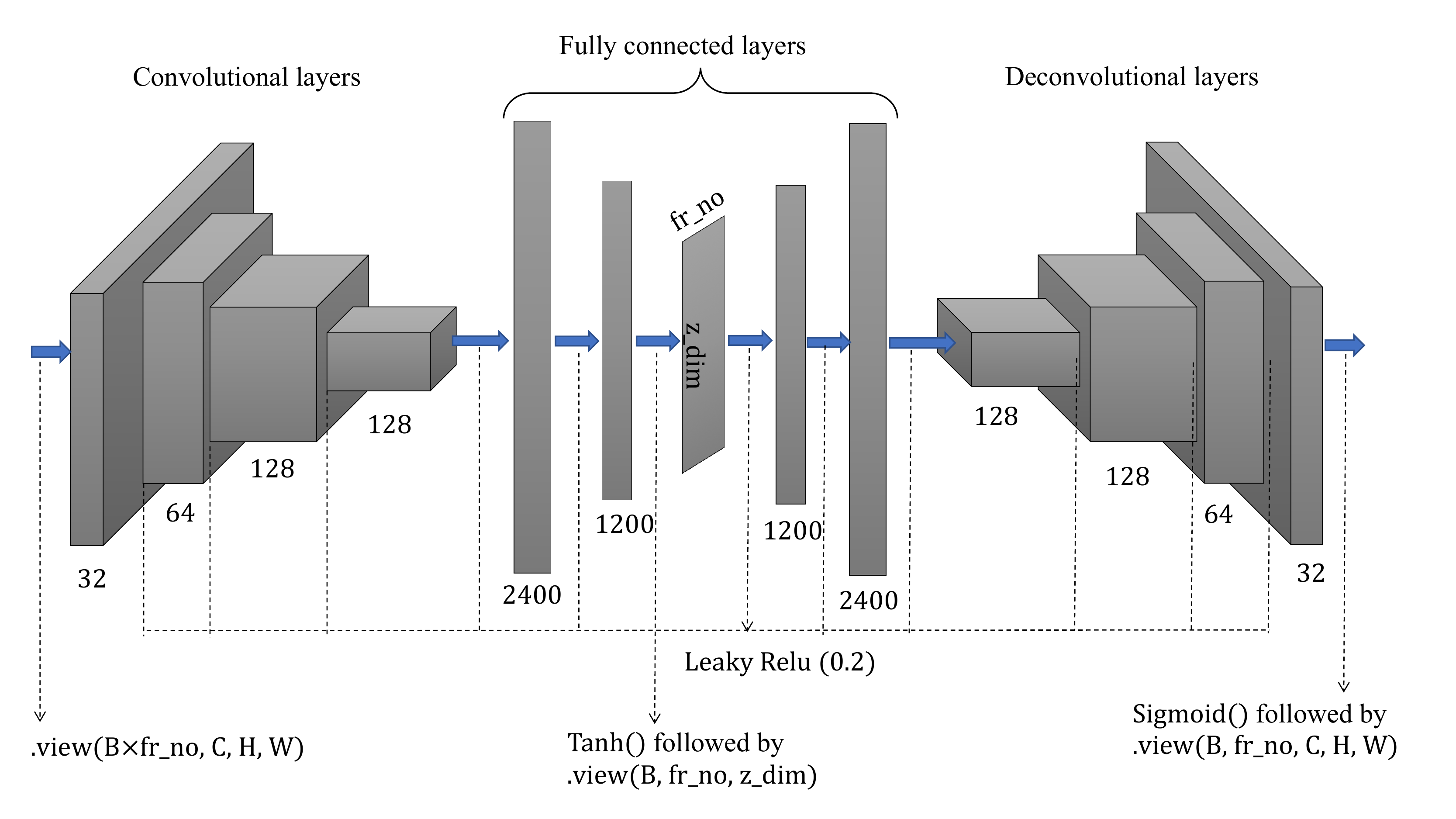}
    \caption{\footnotesize Network architecture of the G-LBM. Input to the network is a batch of video clips. Each video clip is a sequence of consecutive frames with the same background scene. In order to handle the 4D input videos with 2D convolutions, we squeeze the first two axes of input regarding batch and video frames into one axis. We unsqueeze the first dimension where it is necessary.}
    \label{fig:netarch}
    \vspace{-.1in}
\end{figure*}
}
\newcommand{\trainpipeline}{
\begin{figure*}[!ht]
\centering
 \vspace{-.1in}
    \includegraphics[width=0.8\textwidth, trim=0.0in 0.0in 0.0in 0.0in, clip=true]{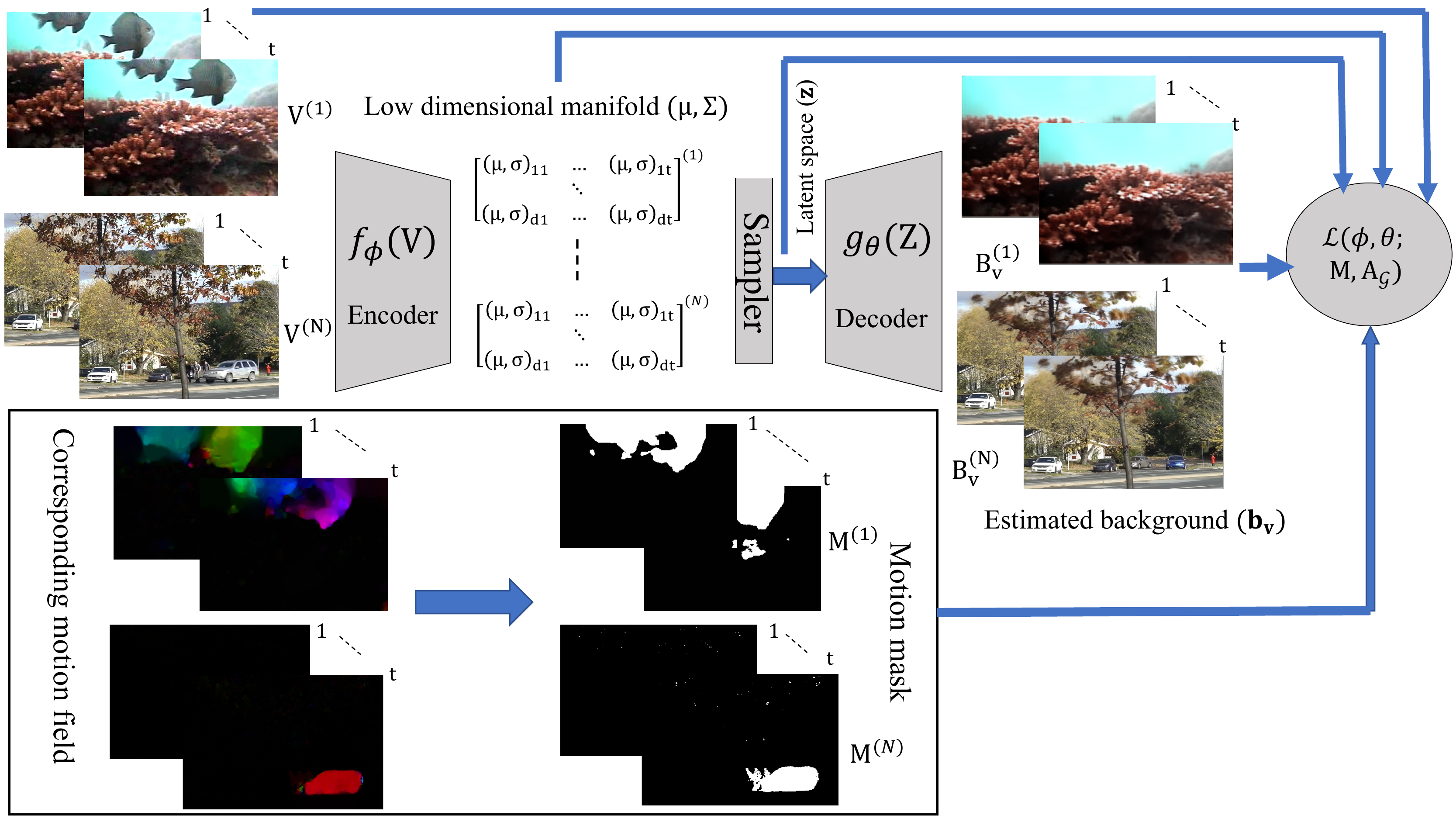}
    \caption{\footnotesize Schematic illustration of the proposed G-LBM training procedure for background model estimation. Given a batch of input video clips $V^{(i)}$ consist of consecutive frames from the same scene to the network, input videos $V^{(i)}$ are mapped through the encoder $f_{\phi}(.)$ to a low-dimensional manifold representing the mean and covariance $(\mu, \Sigma)$ of the latent process $\mathbf{z}$. The latent process $\mathbf{z}$ generates the estimated backgrounds for each video clip through decoder $g_{\theta}(.)$. Imposing the locally linear subspace is done by minimizing the rank of the manifold coordinates correspond to the video frames coming from the same clip ($f_\phi(V^{(i)}):= (\mu, \Sigma)^{(i)}$). Learning the parameters of the non-linear mappings $\phi$, $\theta$ is done by incorporating the reconstruction error between input videos $V^{(i)}$ and estimated background $B^{(i)}_\mathbf{v}$ where motion mask value is zero to the final loss $\mathcal{L}(\phi, \theta; M, A_\mathcal{G})$.}
    \label{fig:trainpip}
    \vspace{-.2in}
\end{figure*}
}
\newcommand{\bmcresult}{
\begin{figure*}[!ht]
\centering
 \vspace{-.1in}
    \includegraphics[width=0.9\textwidth, trim=0.0in 1.2in 0.0in 1.2in, clip=true]{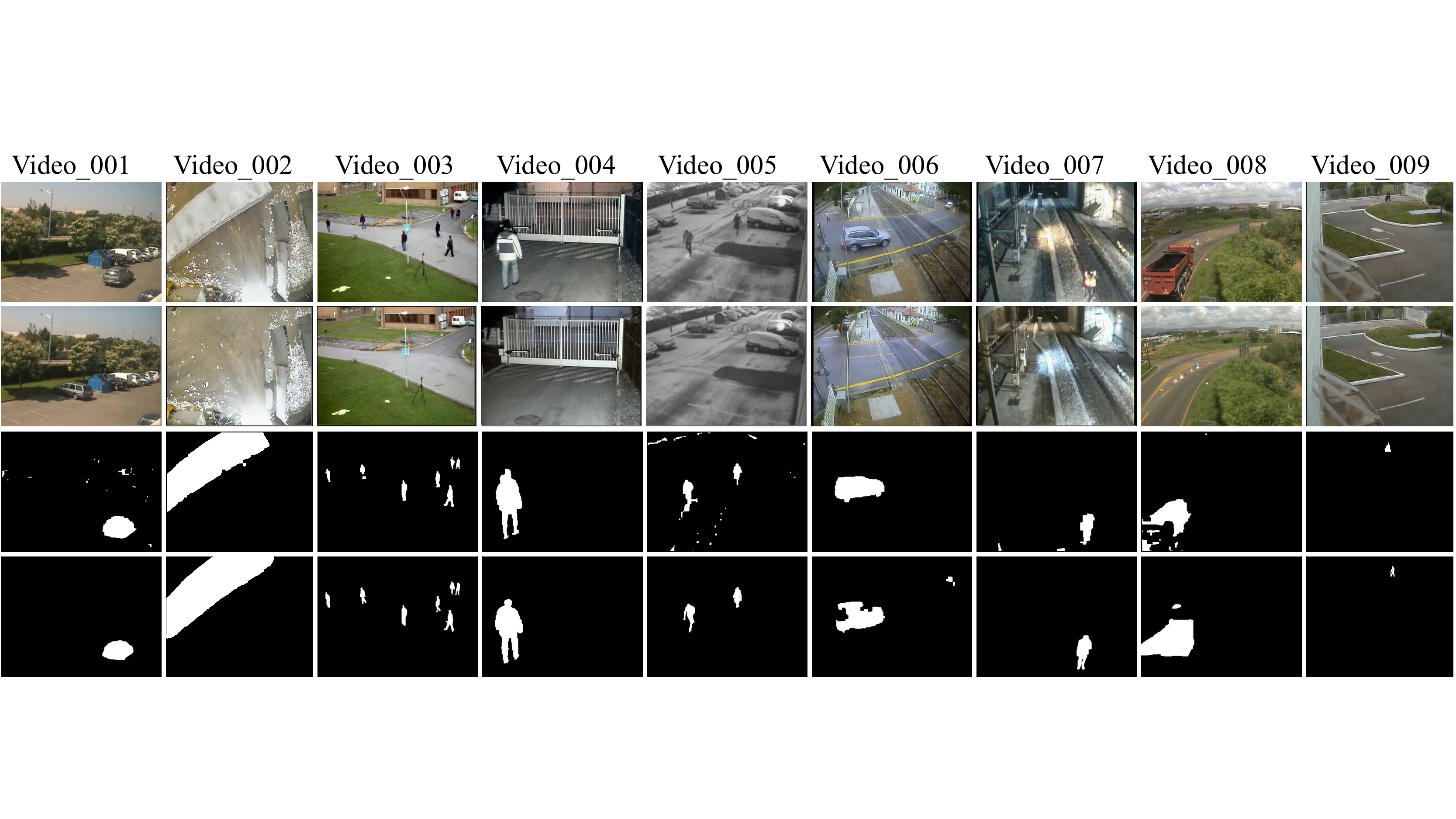}
    \caption{\footnotesize Visual results of the G-LBM over video sequences of BMC2012. First row is the input video frame, second row is the computed background model by G-LBM, third row is the extracted foreground mask by thresholding the difference between input video frame and the G-LBM background model. Last row is the GT foreground mask. }
    \label{fig:bmcresult}
    \vspace{-.2in}
\end{figure*}
}
\newcommand{\sbmCompareResult}{
\begin{figure*}[!ht]
\centering
 \vspace{.1in}
    \includegraphics[width=0.9\textwidth, trim=0.0in 10in 0.0in 8in, clip=true]{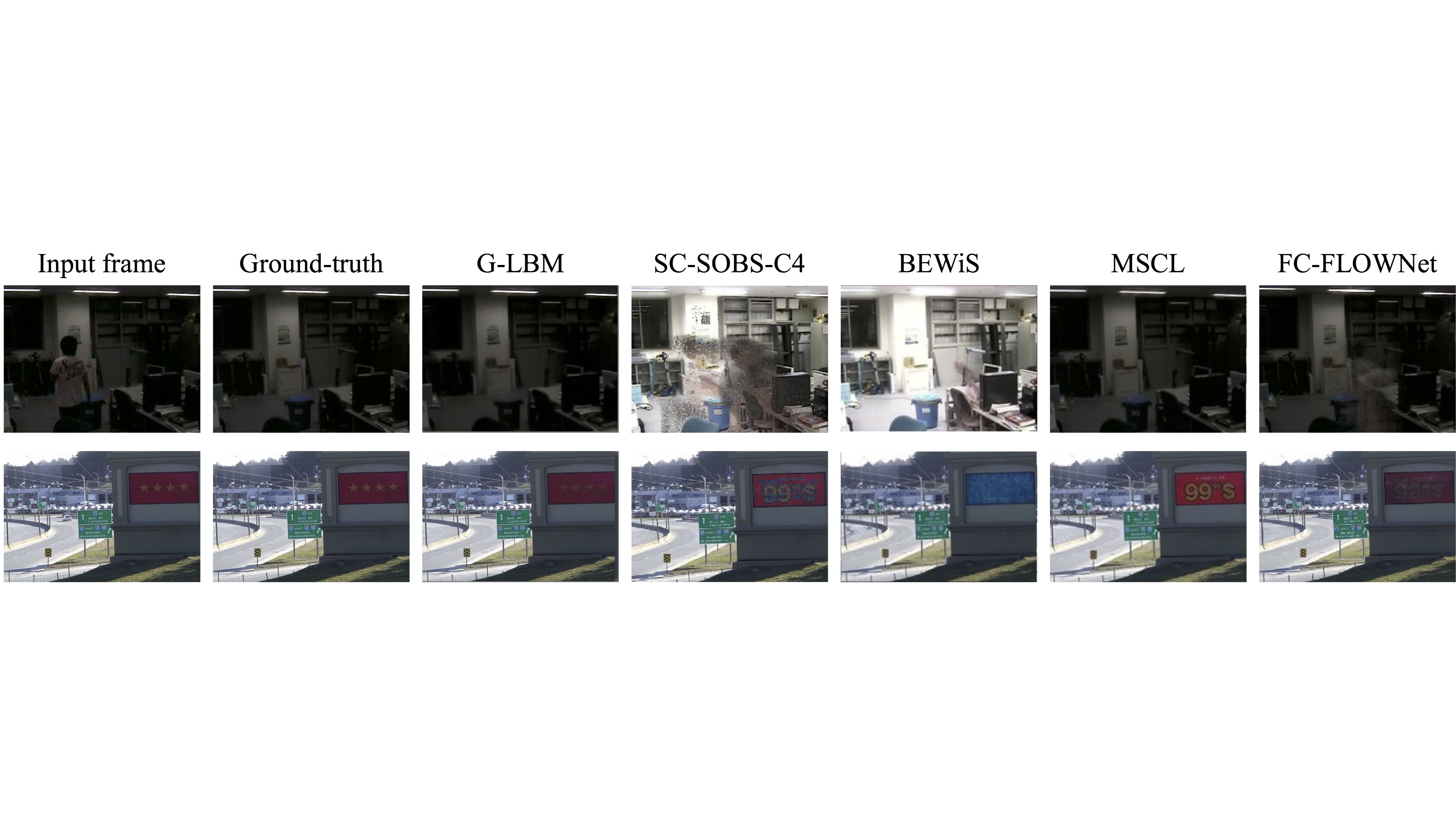}
    \caption{\footnotesize Visual results of the G-LBM compared against other high performing methods on SBMnet-2016 dataset. First and second rows are sample video frames from illumination changes background and motion categories, respectively.}
    \label{fig:sbmCompare}
    \vspace{.1in}
\end{figure*}
\vspace{-.1in}
}
\newcommand{\sbmGlbmGoodVisual}{
\begin{figure*}[!ht]
\centering
 \vspace{-.1in}
    \includegraphics[width=0.9\textwidth, trim=0.0in 1.8in 0.0in 1.8in, clip=true]{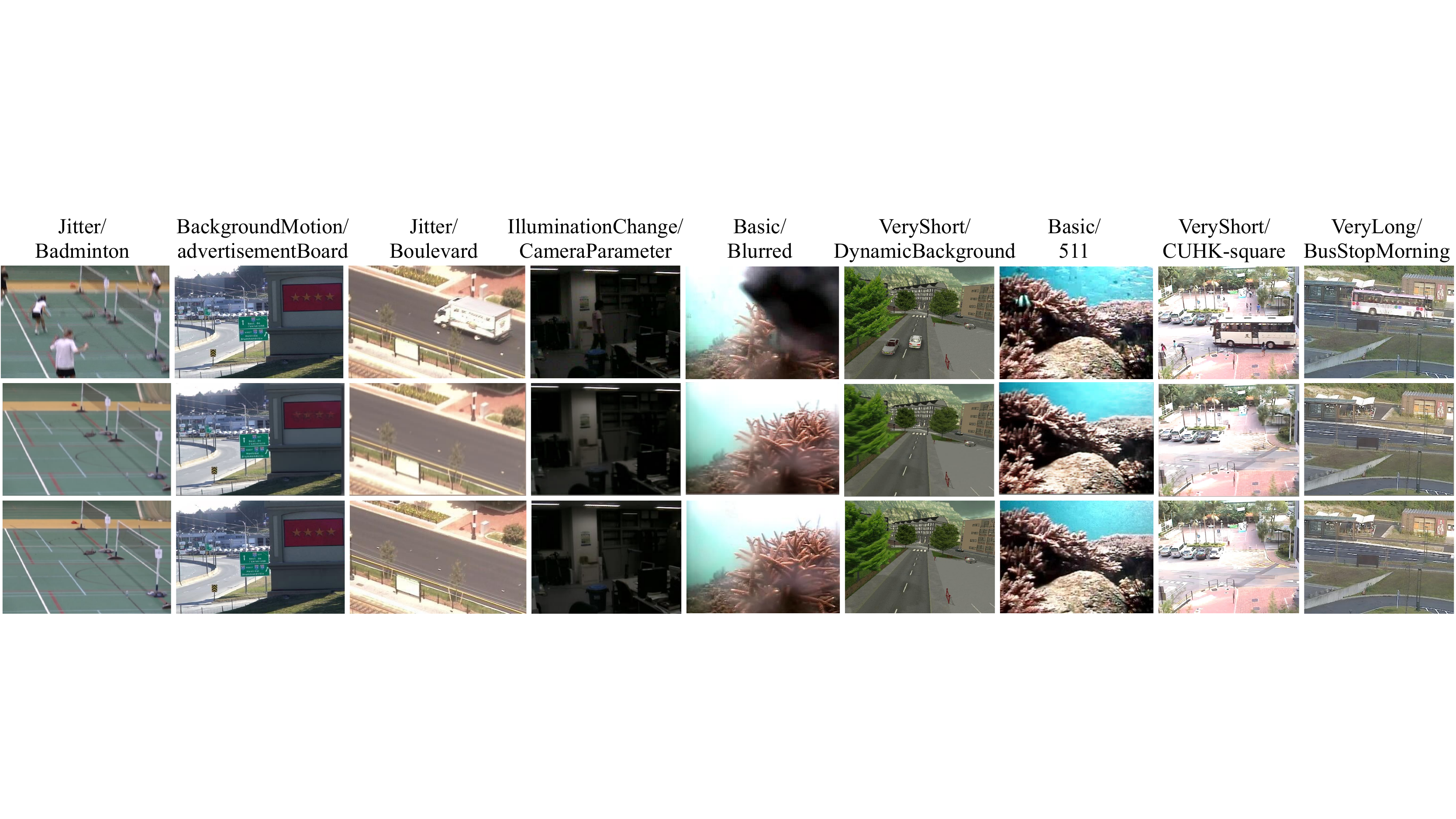}
    \caption{\footnotesize Visual results on categories of SBMnet-2016 that G-LBM successfully models the background with comparable or higher quantitative performance (see \tblref{SBMperCatResult}). First row is the input frame, second row is the G-LBM estimated background, and third row is the GT.}
    \label{fig:sbmGlbmGood}
    \vspace{-.2in}
\end{figure*}
}
\newcommand{\sbmGlbmBadVisual}{
\begin{SCfigure}[1][h]
\centering
    \includegraphics[width=0.5\textwidth, trim=1.0in 0.38in 1.0in 0.5in, clip=true]{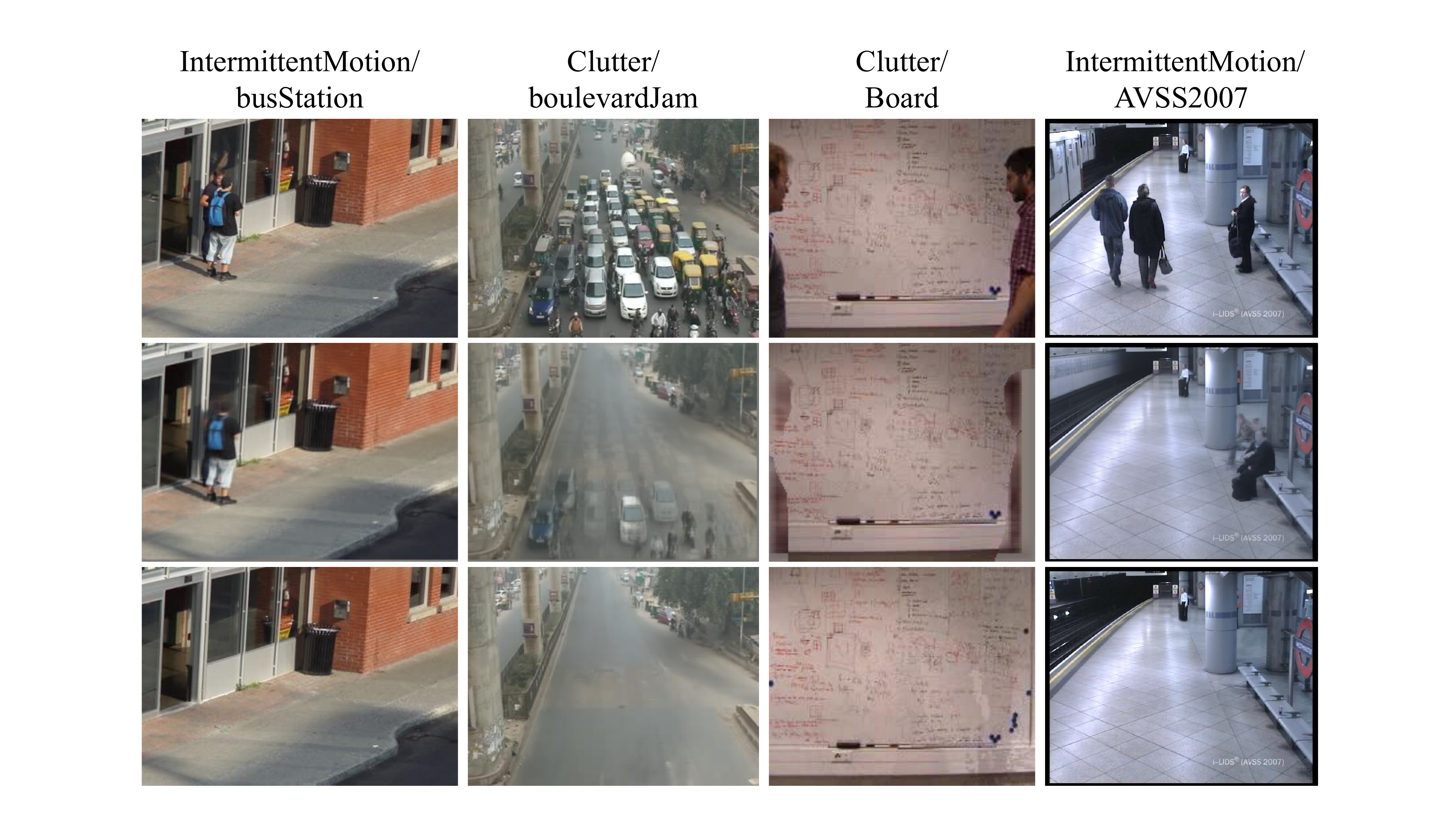}
    \hspace{-.2in}
    \caption{\footnotesize Visualization of G-LBM failure to estimate an accurate model of the background. First row is the input frame, second row is the G-LBM estimated background, and third row is the GT. Since G-LBM is a scene non-specific method it is outperformed by other models that have more specific designs for these challenges.}
    \label{fig:sbmGlbmBad}
    \vspace{-.2in}
\end{SCfigure}
}
\newcommand{\tablresultBMC}{
\begin{table}[t]  
\scriptsize
\begin{center}  
\caption{\footnotesize Comparison of average $F_1$-score on each video of BMC2012 dataset. Long videos are highlighted in gray.}
 \begin{tabular}{ c  c  c  c c c c c c c c c c } 
 Video   &  3TD &  DP-GMM & LSD  & TVRPCA & SRPCA & RMAMR & LR-FSO & GFL & MSCL & \textbf{G-LBM}   \\ 
 \hline
  \rowcolor[gray]{0.8}
 001 & 0.79& 0.72& 0.79& 0.76& 0.79& 0.78& 0.71& 0.78& \textbf{0.80}& 0.73 \\ 
 002 & 0.76& 0.69& 0.80& 0.67& 0.74& 0.71& 0.66& 0.74& 0.78& \textbf{0.85}\\
 003 & 0.70& 0.75& 0.94& 0.68& 0.83& 0.78& 0.70& 0.61&\textbf{0.96}& 0.93 \\ 
 004 & 0.83& 0.80& 0.88& 0.82& 0.81& 0.79& 0.72& 0.88& 0.86& \textbf{0.91}\\
  \rowcolor[gray]{0.8}
 005 & 0.79& 0.71& 0.73& 0.77& 0.80& 0.76& 0.66& \textbf{0.80}& 0.79& 0.71\\
 006 & 0.82& 0.68& 0.80& 0.69& 0.69& 0.65& 0.78& 0.74& 0.74& \textbf{0.85}\\
 007 & 0.73& 0.65& \textbf{0.81}& 0.71& 0.70& 0.64& 0.54& 0.69& 0.76& 0.70\\
 008 & 0.81& 0.78& 0.84& 0.79& 0.84& 0.80& 0.80& 0.81& \textbf{0.89}& 0.76\\
 \rowcolor[gray]{0.8}
 009 & 0.85& 0.79& \textbf{0.92}& 0.88& 0.86& 0.82& 0.82& 0.83& 0.86& 0.69\\
 \hline
 Average &0.78 & 0.73& \textbf{0.83}& 0.75& 0.78& 0.74& 0.71& 0.76& 0.82& 0.79 \\
 \hline
\end{tabular}
\label{tbl:BMCresults}
\end{center}
\vspace{-.3in}
\end{table}
}
\newcommand{\tablresultSBMoverall}{
\addtolength{\tabcolsep}{2pt} 
\begin{table}[t]  
\scriptsize
\begin{center}  
\caption{\footnotesize Comparison of overall model performance evaluated on SBMnet-2016 dataset in terms of metrics averaged over all categories.}
 \begin{tabular}{ c | c  c  c  c  c  c  c  } 
 \footnotesize
Method & Average ranking& AGE& pEPs& pCEPS& MSSSIM& PSNR& CQM \\
\hline
\multicolumn{8}{c}{non-DNN methods} \\
\hline
MSCL &\textbf{1.67} & \textbf{5.9547}& \textbf{0.0524}& \textbf{0.0171}& 0.9410& \textbf{30.8952}& \textbf{31.7049}\\
SPMD &1.83 & 6.0985& 0.0487& 0.0154& \textbf{0.9412}& 29.8439& 30.6499\\
LaBGen-OF &3.00 & 6.1897& 0.0566& 0.0232& 0.9412& 29.8957& 30.7006\\
FSBE &4.33& 6.6204& 0.0605& 0.0217& 0.9373& 29.3378& 30.1777\\
NExBI &9.33 & 6.7778& 0.0671& 0.0227& 0.9196& 27.9944& 28.8810\\
\hline
\multicolumn{8}{c}{DNN-based methods }\\
\hline
\textbf{G-LBM} & 9.67 & 6.8779& 0.0759& 0.0321& 0.9181& \textbf{28.9336}& \textbf{29.7641}\\
BEWiS &\textbf{6.50} & \textbf{6.7094}& \textbf{0.0592}& 0.0266& \textbf{0.9282}& 28.7728& 29.6342\\
SC-SOBS-C4 &12.17 & 7.5183& 0.0711& \textbf{0.0242}& 0.9160& 27.6533& 28.5601\\
FC-FLOWNet &19.83 & 9.1131& 0.1128& 0.0599& 0.9162& 26.9559& 27.8767\\
BE-AAPSA &16.50 & 7.9086& 0.0873& 0.0447& 0.9127& 27.0714& 27.9811\\
 \hline
\end{tabular}
\label{tbl:SBMoverallresults}
\end{center}
\vspace{-.2in}
\end{table}
\addtolength{\tabcolsep}{-2pt} 
}
\newcommand{\tablresultSBMperCat}{
\begin{table}[t]  
\scriptsize
\begin{center}  
\caption{\footnotesize Comparison of G-LBM performance against other approaches evaluated on SBMnet-2016 dataset with respect to averaged AGE over videos in each category.}
 \begin{tabular}{ c | c  c  c  c  c  c  c  c } 
Method & Basic& \shortstack{Intermittent\\ motion}& Clutter& Jitter& \shortstack{Illumination\\ changes}& \shortstack{Background\\ motion}& \shortstack{Very\\ long} & \shortstack{Very\\ short}  \\
\hline
\multicolumn{8}{c}{non-DNN methods} \\
\hline
MSCL &\textbf{3.4019}& \textbf{3.9743}& 5.2695& 9.7403& \textbf{4.4319}& 11.2194& \textbf{3.8214}& 5.7790 \\
SPMD &3.8141& 4.1840& 4.5998& 9.8095& 4.4750& \textbf{9.9115}& 6.0926& 5.9017 \\
LaBGen-OF &3.8421& 4.6433& \textbf{4.1821}& \textbf{9.2410}& 8.2200& 10.0698& 4.2856& \textbf{5.0338}\\
FSBE &3.8960& 5.3438& 4.7660& 10.3878& 5.5089& 10.5862& 6.9832& 5.4912\\
NExBI &4.7466& 4.6374& 5.3091& 11.1301& 4.8310& 11.5851& 6.2698& 5.7134\\
\hline
\multicolumn{8}{c}{DNN-based methods }\\
\hline
\textbf{G-LBM} &4.5013& 7.0859& 13.0786& \textbf{9.1154}& \textbf{3.2735}& \textbf{9.1644}& \textbf{2.5819}& 6.2223\\
BEWiS &\textbf{4.0673}& \textbf{4.7798}& \textbf{10.6714}& 9.4156& 5.9048& 9.6776& 3.9652& \textbf{5.1937}\\
SC-SOBS-C4 &4.3598& 6.2583& 15.9385& 10.0232& 10.3591& 10.7280& 6.0638& 5.2953\\
FC-FLOWNet &5.5856& 6.7811& 12.5556& 10.2805& 13.3662& 10.0539& 7.7727& 6.5094\\
BE-AAPSA &5.6842& 6.6997& 12.3049& 10.1994& 7.0447& 9.3755& 3.8745& 8.0857\\
 \hline
\end{tabular}
\label{tbl:SBMperCatResult}
\end{center}
\vspace{-.1in}
\end{table}
}

\newcommand{\eqnref}[1]{Eq.~(\ref{eqn:#1})}
\newcommand{\figref}[1]{Fig.~\ref{fig:#1}}
\newcommand{\tblref}[1]{Table~\ref{tbl:#1}}
\newcommand{\secref}[1]{Section~\ref{sec:#1}}

\begin{document}
% \linenumbers
\pagestyle{headings}
\mainmatter
\def\ECCVSubNumber{1557}  % Insert your submission number here

%%%%%%%%% TITLE
% \title{Synthesizing Guided 3D Human Pose Estimation via Task Generalization}
% \title{3D Human Pose Estimation via Cross Domain Collaboration}

\title{G-LBM:Generative Low-dimensional Background Model Estimation from Video Sequences}

\begin{comment}
\titlerunning{ECCV-20 submission ID \ECCVSubNumber} 
\authorrunning{ECCV-20 submission ID \ECCVSubNumber} 
\author{Anonymous ECCV submission}
\institute{Paper ID \ECCVSubNumber}
\end{comment}
%******************

%\begin{comment}
\titlerunning{G-LBM}
\author{Behnaz Rezaei \and
Amirreza Farnoosh \and
Sarah Ostadabbas}
\authorrunning{B. Rezaei et al.}
\institute{Augmented Cognition Lab, Electrical and Computer Engineering Department,\\
	Northeastern University, Boston, USA\\
\email{\{brezaei,afarnoosh,ostadabbas\}@ece.neu.edu}\\
\url{http://www.northeastern.edu/ostadabbas/}}
%\end{comment}
%******************

\maketitle

\begin{abstract}
In this paper, we propose a computationally tractable and theoretically supported non-linear low-dimensional generative model to represent real-world data in the presence of noise and sparse outliers. The non-linear low-dimensional manifold discovery of data is done through describing a joint distribution over observations, and their low-dimensional representations (i.e. manifold coordinates). Our model, called generative low-dimensional background model (G-LBM) admits variational operations on the distribution of the manifold coordinates and simultaneously generates a low-rank structure of the latent manifold given the data. Therefore, our probabilistic model contains the intuition of the non-probabilistic low-dimensional manifold learning. G-LBM selects the intrinsic dimensionality of the underling manifold of the observations, and its probabilistic nature models the noise in the observation data. G-LBM has direct application in the background scenes model estimation from video sequences and we have evaluated its performance on SBMnet-2016 and BMC2012 datasets, where it achieved a performance higher or comparable to other state-of-the-art methods while being agnostic to the background scenes in videos. Besides, in challenges such as camera jitter and background motion, G-LBM is able to robustly estimate the background by effectively modeling the uncertainties in video observations in these scenarios\footnote{The code and models are available at: \url{https://github.com/brezaei/G-LBM}.}.

\begin{comment}
We present a probabilistic model for the non-linear low-dimensional manifold discovery of a set of data in the presence of noise and sparse outliers.
 Our model is applied to the context of background model estimation from video sequences which is called generative low-dimensional background model (G-LBM)
\end{comment}
\end{abstract}

\section{Introduction}
    Many high-dimensional real world datasets consist of data points coming from a lower-dimensional manifold corrupted by noise and possibly outliers. In particular, background in videos recorded by a static camera might be generated from a small number of latent processes that all non-linearly affect the recorded video scenes. Linear multivariate analysis such as robust principal component analysis (RPCA) and its variants have long been used to estimate such underlying processes in the presence of noise and/or outliers in the measurements with large data matrices \cite{candes2011robust,vaswani2018robust,javed2018robust}. However, these linear processes may fail to find the low-dimensional structure of the data when the mapping of the data into the latent space is non-linear. For instance background scenes in real-world videos lie on one or more non-linear manifolds, an investigation to this fact is presented in \cite{Javed2017}. Therefore, a robust representation of the data should find the underlying non-linear structure of the real-world data as well as its uncertainties. To this end, we propose a generic probabilistic non-linear model of the background inclusive to different scenes in order to effectively capture the low-dimensional generative process of the background sequences. Our model is inspired by the classical background estimation methods based on the low-dimensional subspace representation enhanced with the Bayesian auto-encoding neural networks in finding the non-linear latent processes of the high-dimensional data. 
Despite the fact that finding the low-dimensional structure of  data has different applications in real world \cite{papadimitriou2000latent,yang2016social,he2017neural}, the main focus of this paper is around the concept of background scene estimation/generation in video sequences. 
%%%%%%%%%%%%%%%%%%%%%%%%%%%%%%%%%%%%%%%%%%%%%%%%%%%%%%%%%%%%%%%%%%%%%
\subsection{Video Background Model Estimation Toward Foreground Segmentation}
Foreground segmentation is the primary task in a wide range of computer vision applications such as moving object detection \cite{rezaei2017background}, video surveillance \cite{bouwmans2014robust}, behavior analysis and video inspection \cite{rezaei2018moving}, and visual object tracking \cite{rezaei2017long}. The objective in foreground segmentation is separating the moving objects from the background which is mainly achieved in three steps of background estimation, background subtraction, and background maintenance.

The first step called background model estimation refers to the extracting a model which describes a scene without foreground objects in a video. In general, a background model is often initialized using the first frame or a set of training frames that either contain or do not contain foreground objects. This background model can be the temporal average or median of the consecutive video frames. However, such models poorly perform in challenging types of environments such as changing lighting conditions, jitters, and occlusions due to the presence of foreground objects. In these scenarios aforementioned simple background models require bootstrapping, and a sophisticated model is then needed to construct the first background image. The algorithms with highest overall performance applied to the SBMnet-2016 dataset, which is the largest public dataset on background modeling with different real world challenges are Motion-assisted Spatio-temporal Clustering of Low-rank (MSCL) \cite{Javed2017}, Superpixel Motion Detector (SPMD) \cite{xu2019robust}, and LaBGen-OF \cite{laugraud2017memoryless}, which are based on RPCA, density based clustering of the motionless superpixels, and the robust estimation of the median, respectively. In practical terms, the main challenge is to obtain the background model when more than half of the training contains foreground objects. This learning process can be achieved offline and thus a batch-type algorithm can be applied. 
Deep neural networks (DNNs) are suitable for this type of tasks and several DNN methods have recently been used in this field. In \secref{relatedwork}, we give an overview of the DNN-based background model estimation algorithms. 

Following the background model estimation, background subtraction in the second step consists of comparing the modeled background image with the current video frames to segment pixels as background or foreground. This is a binary classification task, which can be achieved successfully using a DNN. Different methods for the background subtraction have been developed, and we refer the reader to look at \cite{Bouwmans2019,bouwmans2019background} for comprehensive details on these methods. While we urge the background subtraction process to be unsupervised given the background model, the well performing methods are mostly supervised. The three top algorithms on the CDnet-2014 \cite{wang2014cdnet} which is the large-scale real-world dataset for background subtraction are supervised DNN-based methods, namely different versions of FgSegNet \cite{lim2018foreground} , BSPVGAN \cite{zheng2019novel}, cascaded CNN \cite{wang2017interactive}, followed by three unsupervised approaches, WisennetMD \cite{jiang2017wesambe}, PAWCS \cite{st2015self}, IUTIS \cite{bianco2017combination}.  
%%%%%%%%%%%%%%%%%%%%%%%%%%%%%%%%%%%%%%%%%%%%%%%%%%%%%%%%%%%%%%%%%%%%%%
\subsection{Related Work on Background Model Estimation}
\label{sec:relatedwork}
DNNs have been widely used in modeling the background from video sequences due to their flexibility and power in estimating complex models. Aside from prevalent use of convolutional neural networks (CNNs) in this field, successful methods are mainly designed based on the deep auto-encoder networks (DAE), and generative adversarial networks (GAN).
\begin{enumerate}
    \item Model architectures based on convolutional neural networks (CNNs):
FC-FlowNet model proposed in \cite{halfaoui2016cnn} is a  CNN-based architecture inspired from the FlowNet proposed by Dosovitskiy et al. in \cite{dosovitskiy2015flownet}. FlowNet is a two-stage architecture developed for the prediction of the optical flow motion vectors: A contractive stage, composed of a succession of convolutional layers, and a refinement stage, composed of deconvolutional layers. FC-FlowNet modifies this architecture by creating a fully-concatenated version which combines at each convolutional layer multiple feature maps representing different high level abstractions from previous stages. 
%The main idea of FC-FlowNet is to improve the quality of the feature maps during both contractive and refinement stages by taking advantage of the redundancy gained from other layers. 
Even though FC-FlowNet is able to model background in mild challenges of real-world videos, it fails to address challenges such as clutters, background motions, and illumination changes.

    \item Model architectures based on deep auto-encoding networks (DAEs):
One of the earliest works in background modeling using DAEs was presented in \cite{xu2014dynamic}. Their model is a cascade of two auto-encoder networks. The first network approximates the background images from the input video. Background model is then learned through the second auto-encoder network. Qu et al. \cite{qu2016motion} employed context-encoder to model the motion-based background from a dynamic foreground. Their method aims to restore the overall scene of a video by removing the moving foreground objects and learning the feature of its context. 
%A context-encoder is also applied to inpaint the missing background pixels in the empty regions to generate the final background model.
%An advantage of this method is that the performance of background modeling will not be affected when the camera is moving fast.% 
Both aforementioned works have limited number of experiments to evaluate their model performance.
More recently two other unsupervised models for background modeling inspired by the successful novel auto-encoding architecture of U-net \cite{ronneberger2015u} have been proposed in \cite{tao2017background,mondejar2019end}.
BM-Unet and its augmented version presented by Tao et al. \cite{tao2017background} is a background modelling method based on the U-net architecture. 
%Given a certain frame as input to the network, BM-Unet generates the corresponding probabilistic heat map of the color values as the output background image. 
They augment their baseline model 
%utilising color, intensity differences, and optical flow between a reference and a target frame 
with the aim of robustness to rapid illumination changes and camera jitter. However, they did not evaluate their proposed model on the complete dataset of SBMnet-2016.
DeepPBM in \cite{farnoosh2019deeppbm} is a generative scene-specific background model based on the variational auto-encoders (VAEs) evaluated on the BMC2012 dataset compared with RPCA. 
Mondejar et al. in \cite{mondejar2019end} proposed an architecture for simultaneous background modeling and subtraction consists of two cascaded networks which are trained together. Both sub-networks have the same architecture as U-net architecture. The first network, namely, background model network takes the video frames as input and results in $M$ background model channel as output. The background subtraction sub-network, instead, takes the $M$ background model channels plus the target frame channels from the same scene as input. 
%The final output is a binary segmentation of each pixel for the given target frame corresponding to the background and foreground. 
The whole network is trained in a supervised manner given the ground truth binary segmentation. Their model is scene-specific and cannot be used for unseen videos. 
    \item Model architectures based on generative adversarial network (GAN):
Considering the promising paradigm of GANs for unsupervised learning, they have been used in recent research of background modeling. Sultana et al. in \cite{sultana2019unsupervised} designed an unsupervised deep context prediction (DCP) for background initialization using hybrid GAN. DCP is a scene-specific background model which consists of four steps: (1) Object masking by creating the motion masks. 
%Motion masks are described by estimating the motion via dense optical flow. 
(2) Evaluating the missing regions resulted from masking the motions using the context prediction hybrid GAN. 
%This step is basically an unsupervised visual feature learning. 
(3) improving the fine texture details by scene-specific fine tuning of VGG-16 network pre-trained on ImageNet \cite{deng2009imagenet}. (4) Obtaining the final background model by applying modified Poisson blending technique. Their model is a scene-specific containing two different networks which are trained separately.
\end{enumerate}
%%%%%%%%%%%%%%%%%%%%%%%%%%%%%%%%%%%%%%%%%%%%%%%%%%%%%%%%%%%%%
\subsection{Our Contributions}
Background models are utilized to segment the foreground in videos, generally regarded as object of interest towards further video processing. Therefore, providing a robust background model in various computer vision applications is an essential preliminary task. However, modeling the background in complex real-world scenarios is still challenging due to presence of dynamic backgrounds, jitter induced by unstable camera, occlusion, illumination changes. None of the approaches proposed so far could address all the challenges in their model. Moreover, current background models are mostly scene-specific. Therefore, DNN models need to be retrained adjusting their weights for each particular scene and non-DNN models require parameter tuning for optimal result on different video sequences, which makes them unable to extend to unseen scenes.
According to the aforementioned challenges in the background modeling, we propose our generative low-dimensional background model (G-LBM) estimation approach that is applicable to different video sequences. Our main contributions in this paper are listed as follows:
\begin{itemize}
\item The G-LBM, our proposed background model estimation approach is the first generative probabilistic model which learns the underlying nonlinear manifold of the background in video sequences. 
\item The probabilistic nature of our model yields the uncertainties that correlate well with empirical errors in real world videos, yet maintains the predictive power of the deterministic counter part. This is verified with the extensive experiments on videos with camera jitters, background motions, and long videos.

\item  The G-LBM is scene non-specific and can be extended to the new videos with different scenes. 

\item We evaluated the proposed G-LBM on large scale background model datasets SBMnet as well as BMC. Experiments show promising results for modeling the background under various challenges.
\end{itemize}
 
 Our contributions are built upon the assumption that there is a low-dimensional  non-linear latent space process that  generates background in different videos. In addition, background from the videos with the same scene can be non-linearly mapped into a lower dimensional subspace. In different words the underlying non-linear manifold of the background in different videos is locally linear.

%The rest of the paper is organized as follows. In \secref{method}, we introduce our generative model, G-LBM for which we drive the variational inference method and present our background model estimation using the VAE structure. In \secref{experiment}, we demonstrate the performance of our approach on background initialization/modeling problem in computer vision. Finally, we conclude the paper in \secref{conclusion}.

\section{Generative Low-dimensional Background Model}
    \label{sec:method}
We designed an end-to-end architecture that performs a scene non-specific background model estimation by finding the low-dimensional latent process which generates the background scenes in video sequences. An overview of our generative model is presented in \figref{grmodel}. As described in \secref{inferencemodel}, the latent process $\mathbf{z}$ is estimated through a non-linear mapping from the corrupted/noisy observations $\mathbf{v}=\{\mathbf{v_1}, ..., \mathbf{v_n}\}$ parameterized by $\phi$.
\graphicalmodel

\textit{Notation}: In the following, a diagonal matrix with entries taken from vector $\mathbf{x}$ is shown as $diag(\mathbf{x})$. Vector of $n$ ones is shown as $\mathbf{1}_n$ and $n \times n$ identity matrix is $I_n$.
The Nuclear norm of a matrix $B$ is $||B||_\star$ and its $l_1$-norm is $||B||_1$. The Kronecker product of matrices $A$ and $B$ is $A \otimes B$. Khatri–Rao product is defined as
$\displaystyle{A \ast B =\left(A _{ij}\otimes B _{ij}\right)_{ij}}$, in which the $ij$-th block is the $m_ip_i \times n_jq_j$ sized Kronecker product of the corresponding blocks of $A$ and $B$, assuming the number of row and column partitions of both matrices is equal. 

%%%%%%%%%%%%%%%%%%%%%%%%%%%%%%%%%%%%%%%%%%%%%%%%%%%%%%%%%%%%%%%%%%%%%%%%%%%%%%%%%%%%%%%%%%
\subsection{Nonlinear Latent Variable Modeling of Background in Videos}
\label{sec:inferencemodel}
\textbf{Problem formulation:} Suppose that we have $n$ data points $\{\mathbf{v}_1, \mathbf{v}_2, ..., \mathbf{v}_n\} \subset \mathbb{R}^m$, and a graph $\mathcal{G}$ with  $n$ nodes  corresponding to each data point with the edge set
$\mathcal{E_G} = \{(i, j) | \mathbf{v}_i ~ \text{and}~ \mathbf{v}_j ~ \text{are neighbours}\}$. In context of modeling the background $\mathbf{v}_i$ and $\mathbf{v}_j$ are neighbours if they are video frames from the same scene. We assume that there is a low-dimensional (latent) representation of the high-dimensional data $\{\mathbf{v}_1, \mathbf{v}_2, ..., \mathbf{v}_n\}$ with coordinates $\{\mathbf{z}_1,\mathbf{z}_2, ..., \mathbf{z}_n\} \subset \mathbb{R}^d$, where $d \ll m$. It is helpful to concatenate data points from the same clique in the graph to form $V^{(i)}$ and all the cliques to form the $\mathbf{V}=concat(V^{(1)}, ..., V^{(N)})$.\\
\textbf{Assumptions:} Our essential assumptions are as follows: (1) Latent space is locally linear in a sense that neighbour data points in the graph $\mathcal{G}$ lie in a lower dimensional subspace. On the other hand, mapped neighbour data points in latent space $\{(\mathbf{z}_i, \mathbf{z}_j)|(i,j) \in \mathcal{E_G}\}$ belong to a subspace with dimension lower than the manifold dimension. (2) Measurement dataset is corrupted by sparse outliers (foreground in the videos). 

Under these assumptions and given the graph $\mathcal{G}$, we aim to find the non-linear mapping from observed input data into a low-dimensional latent manifold and the distribution over the latent process $\mathbf{z}$, $p(\mathbf{z}|\mathcal{G},\mathbf{v})$, which best describes the data such that the samples of the estimated latent distribution can generate the data through a non-linear mapping.
In the following, we describe the main components of our Generative Low-dimensional Background Model (G-LBM).

\textit{Adjacency and Laplacian matrices}: The edge set of $\mathcal{G}$ for $n$ data points specifies a $n \times n$ symmetric adjacency matrix $A_\mathcal{G}$. $a_{ij}$ defined as $i, j\text{th}$ element of $A_\mathcal{G}$, is $1$ if $\mathbf{v}_i$ and $\mathbf{v}_j$ are neighbours and $0$ if not or $i=j$ (diagonal elements). Accordingly, the Laplacian matrix is defined as:
$L_\mathcal{G} = diag(A_\mathcal{G}\mathbf{1}_n)-A_\mathcal{G}$.

\textit{Prior distribution over $\mathbf{z}$}: We assume that the prior on the latent variables $\mathbf{z}_i$, $i\in \{1, ..., n\}$ is a unit variant Gaussian distribution $\mathcal{N}(\mathbf{0}, I_{d})$. This prior as a multivariate normal distribution on concatenated $\mathbf{z}$ can be written as:
\begin{align} \label{eqn:prior}
\begin{split}
    p(\mathbf{z}|A_\mathcal{G})=\mathcal{N}(\mathbf{0}, \Sigma), ~~~
    \text{where} ~~~ \Sigma^{-1} = 2 L_\mathcal{G} \otimes I_{d}.  
\end{split}  
\end{align}

\textit{Posterior distribution over $\mathbf{z}$}: Under the locally linear dependency assumption on the latent manifold, the posterior is defined as a multivariate Gaussian distribution given by \eqnref{posterior}. Manifold coordinates construct the expected value $\Lambda$ and covariance $\Pi$ of the latent process variables corresponding to the neighbouring high dimensional points in graph $\mathcal{G}$.
\begin{align} \label{eqn:posterior}
\scriptsize
& p(\mathbf{z}|A_\mathcal{G}, \mathbf{v}) =   \mathcal{N}(\Lambda, \Pi),~~~ \text{where}\\ \nonumber
& \Pi^{-1} = 2 L_\mathcal{G} \ast  [diag(f_\phi^\sigma(\mathbf{v}_1)), \dots, diag(f_\phi^\sigma(\mathbf{v}_n))]^T  [diag(f_\phi^\sigma(\mathbf{v}_1)), \dots, diag(f_\phi^\sigma(\mathbf{v}_n))] \\ \nonumber
& \Lambda = [ f_\phi^\mu(\mathbf{v}_1)^T,\dots, f_\phi^\mu(\mathbf{v}_n)^T]^T \in \mathbb{R}^{nd},
\end{align}
where $f_\phi^\sigma(\mathbf{v}_i)$ and $f_\phi^\mu(\mathbf{v}_i), \,\, \text{for} \,\, i=\{1,\dots, n\}$ are corresponding points on the latent manifold mapped from high-dimensional point $\mathbf{v}_i$ by nonlinear function $f_\phi(.)$. These points are treated as the mean and variance of the latent process respectively. Our aim is to infer the latent variables $\mathbf{z}$ as well as the non-linear mapping parameters $\phi$ in G-LBM. We infer the parameters by minimizing the reconstruction error when generating the original data points through mapping the corresponding samples of the latent space into the original high-dimensional space. Further details on finding the parameters of the non-linear mapping in G-LBM from video sequences are provided in \secref{VAEmodel}. 
%%%%%%%%%%%%%%%%%%%%%%%%%%%%%%%%%%%%%%%%%%%%%%%%%%%%%%%%%%%%%%%%%%%%%%%%%%%%%%%
\subsection{Background Model Estimation in G-LBM using VAE}
\label{sec:VAEmodel}
\trainpipeline
Consider that backgrounds in video frames $\mathbf{v}$ belong to $ \{\mathbf{v}_1,\dots,\mathbf{v}_n\}$, each of size $m=w\times h$ pixels, are generated from $n$ underlying probabilistic latent process vectorized in $\mathbf{z}\in \mathbb{R}^d$ for $d \ll m$. Video frame $\mathbf{v}_i$ is interpreted as the corrupted background in higher dimension by sparse perturbations/outliers called foreground objects, and vector $\mathbf{z}_i$ is interpreted as the low-dimensional representation of the background in video frame $\mathbf{v}_i$. The neighbourhood graph $\mathcal{G}$ represents the video frames recorded from the same scene as nodes of a clique in the graph. A variational auto-encoding considers the joint probability of the background in input video frames $\mathbf{v}$ and its representation $\mathbf{z}$ to define the underlying generative model as $p_\theta(\mathbf{v}, \mathbf{z}|A_\mathcal{G}) = p_\theta(\mathbf{v} \vert \mathbf{z}) p(\mathbf{z}|A_\mathcal{G})$, where $p(\mathbf{z}|A_\mathcal{G})$ is the Gaussian prior for latent variables $\mathbf{z}$ defined in \eqnref{prior}, and $p_\theta(\mathbf{v}|\mathbf{z})$ is the generative process of the model illustrated in \figref{grmodel}. 

In variational auto-encoding (VAE), the generative process is implemented by the decoder part $g_\theta(.)$ that is parameterized by a DNN with parameters $\theta$ as demonstrated in \figref{trainpip}. 
In the encoder part of the VAE, the posterior distribution is approximated with a variational posterior $p_\phi(\mathbf{z}\vert\mathbf{v}, A_\mathcal{G})$ defined in \eqnref{posterior} with parameters $\phi$ corresponding to non-linear mapping $f_\phi(.)$ which is also parameterized by a DNN. 
We assume that backgrounds in input video frames specified as $\mathbf{b}_\mathbf{v}$ belong to $\{\mathbf{b}_\mathbf{v}^1, \dots, \mathbf{b}_\mathbf{v}^n \}$ are generated by $n$ underlying process specified as $\mathbf{z}$ belong to $\{\mathbf{z}_1, \dots, \mathbf{z}_n \}$ in \figref{grmodel}. It is also helpful to concatenate backgrounds from the same video clique in the graph to form $B_\mathbf{v}^{(i)}$ and all the background cliques to form the $\mathbf{B_v}=concat(B_\mathbf{v}^{(1)}, ..., B_\mathbf{v}^{(N)})$.

The efforts in inferring the latent variables $\mathbf{z}$ as well as the parameters $\phi$ in G-LBM results in maximization of the lower bound $\mathcal{L}$ of the marginal likelihood of the background in video observations. \cite{2013arXiv1312.6114K,2016arXiv160100670B}.
Therefore, the total VAE objective for the entire video frames becomes:
\begin{align} 
\label{eqn:vaeObjective}
    & \log p(\mathbf{v}|A_\mathcal{G}, \phi) \geq \\ \nonumber
    & \mathbb{E}_{q_\phi(\mathbf{z}|\mathbf{v}, A_\mathcal{G})}\big[\log p_\theta(\mathbf{v}|\mathbf{z})\big]- KL\big(q_\phi(\mathbf{z}|\mathbf{v}, A_\mathcal{G})||p(\mathbf{z}|A_\mathcal{G})\big) :=\mathcal{L}(p(\mathbf{z|A_\mathcal{G}}), \phi, \theta).
\end{align}

The first term in \eqnref{vaeObjective} can be interpreted as the negative reconstruction error, which encourages the decoder to learn to reconstruct the original input. The second term is the Kullback-Leibler (KL) divergence between prior defined in \eqnref{prior} and variational posterior of latent process variables defined in \eqnref{posterior}, which acts a regularizer to penalize the model complexity. The expectation is taken with respect to the encoder’s distribution over the representations given the neighborhood graph adjacency and input video frames. The KL term can be calculated analytically in the case of Gaussian distributions, as indicated by \cite{2013arXiv1312.6114K}. 

In proposed G-LBM model, the VAE objective is further constrained to linear dependency of the latent variables corresponding to the input video frames of the same scene (neighbouring data points in graph $\mathcal{G}$). This constraint is imposed by minimizing the rank of the latent manifold coordinates mapped from the video frames in the same clique in the graph $\mathcal{G}$.
\begin{align}
%\scriptsize
\label{eqn:rank}
\begin{split}
rank(f_\phi(V^{(i)})) < \delta ~~~\forall i\in\{1, ..., N\}, 
\end{split}
\end{align}
where  $f_\phi(V^{(i)})$ is the estimated mean and variance of the latent process $\mathbf{z}$ relative to the concatenated input video frames $V^{(i)}$ coming from the same scene (clique in the $\mathcal{G}$). $N$ is the total number of cliques. As schematic illustration presented in \figref{trainpip} s. For our purpose of background modeling in G-LBM, given the knowledge that sparse ouliers to the backgrounds in the videos are moving objects, we extract a motion mask from the video frames. This motion mask is incorporated into the reconstruction loss of the VAE objective in \eqnref{vaeObjective} to provide a motion aware reconstruction loss in G-LBM. Given the motion mask $M$ the VAE objective is updated as follows.
\begin{align}
\label{eqn:vaeObjnew}
    & \mathcal{L}(p(\mathbf{z}|A_\mathcal{G}), \phi, \theta; M) = \\ \nonumber
    & \mathbb{E}_{q_\phi(\mathbf{z}|\mathbf{v}, A_\mathcal{G})}\big[\log p_\theta(\mathbf{v}|\mathbf{z}, M)\big]- KL\big(q_\phi(\mathbf{z}|\mathbf{v}, A_\mathcal{G})||p(\mathbf{z}|A_\mathcal{G})\big). 
\end{align}
VAE tries to minimize the reconstruction error between input video frames and estimated backgrounds where there is no foreground object (outlier) given by the motion mask. This minimization is done under an extra constraint to the VAE objective which imposes the sparsity of the outliers/perturbations defined as:
\begin{align}
%\scriptsize
\label{eqn:sparsity}
\begin{split}
\parallel M^{(i)}(V^{(i)} - B_\mathbf{v}^{(i)})\parallel_{0} < \epsilon ~~~\forall i\in\{1, ..., N\}, 
\end{split}
\end{align}
where $\parallel.\parallel_{0}$ is the $l_0$-norm of the difference between concatenated input observations $V^{(i)}$ from each scene and their reconstructed background $B_\mathbf{v}^{(i)} = g_\phi(Z^{(i)})$ where motion mask $M$ is present. Putting objective function and constrains  together the final optimization problem to train the G-LBM model becomes: 
\begin{align}
\label{eqn:vaeObjnew}
    & \min~ \mathcal{L}(p(\mathbf{z}|A_\mathcal{G}), \phi, \theta; M)   \\ \nonumber
    &\text{st:}~~~M^{(i)}(V^{(i)} - B_\mathbf{v}^{(i)})\parallel_{0} < \epsilon
    ~~~\text{and} ~~~ rank(f_\phi(V^{(i)})) < \delta ~~~\forall i\in\{1, ..., N\}.
\end{align}
In order to construct the final loss function to be utilized in learning the parameters of the encoder and decoder in G-LBM model, we used Nuclear norm $\parallel . \parallel_\star$ given by the sum of singular values and $l_1$-norm $\parallel.\parallel_{1}$ as the tightest convex relaxations of the $rank(.)$ and $l_0$-norm respectively. Substituting the reconstruction loss and analytical expression of the KL term in \eqnref{vaeObjnew} the final loss of G-LBM to be minimized is:
\begin{align}
\label{eqn:finalLoss}
 & \mathcal{L}(\phi, \theta;M, A_\mathcal{G})= \\ \nonumber 
 & \sum_{i=1}^{N}BCE(\bar{M}^{(i)}V^{(i)}, B^{(i)})- \frac{1}{2} \big( tr(\Sigma^{-1} \Pi - I) + \Lambda^T \Sigma^{-1} \Lambda + log\frac{\vert\Sigma\vert}{\vert\Pi\vert}
 \big)+ \\ \nonumber
 & \beta \sum_{i=1}^{N}\parallel M^{(i)}(V^{(i)}-B^{{(i)}})\parallel_1 + \alpha 
 \sum_{i=1}^{N}tr(\sqrt{f_\phi(V^{(i)})^T f_\phi(V^{(i)})}).
\end{align}

The motion mask in our reconstruction loss is constructed by computing the motion fields using the coarse2fine optical flow \cite{pathak2017learning} between each pair of consecutive frames in the given sequence of video frames $V^{(i)}$ from the same scene. Using the motion information we compute a motion mask $M^{(i)}$. Let $\mathbf{v}_i$ and $\mathbf{v}_{i-1}$ be the two consecutive frames in $V^{(i)}$ and $h^x_{i,k}$ and $h^y_{i,k}$ be the horizontal and vertical component of motion vector $\mathbf{m}_i$ at position $k$ computed between frames $\mathbf{v}_i$ and $\mathbf{v}_{i-1}$ respectively. $\mathbf{m}_i \in \{0, 1\}$ is the corresponding vectored motion mask computed as:
\begin{align}
\label{eqn:motionMask}
    m_{i,k}= \left \{
        \begin{array}{lr}
             1,~~~ & \text{if }~ \sqrt{(h^x_{i,k})^2+(h^y_{i,k})^2} < \tau   \\
             0, &  \text{otherwise,}
        \end{array}
        \right.
\end{align}
where threshold of motion magnitude $\tau$ is selected adaptively as a factor of the average of all pixels in motion field such that all pixels in $V^{(i)}$ exhibiting motion larger than $\tau$ definitely belong to the foreground not the noise in the background in videos. By concatenating all the motion vectors $\mathbf{m}_i$ computed from input $V^{(i)}$ we construct the $M^{(i)}$. Concatenation of all $M^{(i)}$ is specified as $\mathbf{M}$.
%%%%%%%%%%%%%%%%%%%%%%%%%%%%%%%%%%%%%%%%%%%%%%%%%%%%%%%%%%
\subsection{G-LBM Model Architecture and Training Setup}
\label{sec:architecture}
\networkarchitecture
The encoder and decoder parts of the VAE are both implemented using CNN architectures specified in \figref{netarch}. The encoder takes the video frames as input and outputs the mean and variance of the distribution over underlying low-dimensional latent process. The decoder takes samples drawn from latent distributions as input and outputs the recovered version of the background in original input. 
We trained G-LBM using the VAE architecture in \figref{netarch} by minimizing the loss function defined in \eqnref{finalLoss}. We trained the G-LBM model using gradient descent based on Adam optimization with respect to the parameters of the encoder and decoder, i.e., $\theta$ and $\phi$, respectively. We employed learning rate scheduling and gradient clipping in the optimization setup. Training was performed on batches of size 3 video clips with 40 consecutive frames i.e 120 video frames in every input batch for 500 epochs.

\section{Experimental Results}
    \label{sec:experiment}
In this section, the performance of the proposed G-LBM is evaluated on two publicly available datasets BMC2012 and SBMnet-2016 \cite{vacavant2012benchmark,jodoin2017extensive}. Both quantitative and qualitative performances compared against state-of-the-art methods are provided in \secref{bmc} and \secref{sbmnet}. Results show comparable or better performance against other state-of-the-art methods in background modeling.
\subsection{BMC2012 Dataset}
\label{sec:bmc}
We evaluated the performance of our proposed method on the BMC2012 benchmark dataset \cite{vacavant2012benchmark}. We used $9$ real-world surveillance videos in this dataset, along with encrypted ground truth (GT) masks of the foreground for evaluations. This dataset focuses on outdoor situations with various weather and illumination conditions making it suitable for performance evaluation of background subtraction (BS) methods in challenging conditions. The evaluation metrics are computed by the software that is provided with the dataset. Since this dataset is designed for the BS task in order to be able to do comparison on this dataset, we further performed BS by utilizing the output of the trained G-LBM model. 

To extract the masks of the moving objects in videos, we first trained our model G-LBM using all of the video frames of short videos (with less than 2000 frames) and first 10000 frames of the long videos as explained in \secref{architecture}. After the model was trained, we fed the same frames to the network to estimate the background for each individual frame. Finally, we used the estimated background of each frame to find the mask of the moving objects by thresholding the difference between the original input frame and the estimated background. \tblref{BMCresults} shows the quantitative performance of G-LBM compared to other BS methods including 3TD\cite{oreifej2012simultaneous}, DP-GMM\cite{haines2013background}, LSD\cite{liu2015background}, TVRPCA\cite{cao2015total}, SRPCA\cite{javed2016spatiotemporal}, RMAMR\cite{ortego2016rejection}, LR-FSO\cite{xue2013foreground}, GFL\cite{xin2015background}, MSCL\cite{Javed2017}. \figref{bmcresult} shows the estimated backgrounds and extracted masks by G-LBM model on sample video frames in the BMC2012 dataset. Considering that G-LBM is a scene non-specific model of the background and the task of BS is performed by simply thresholding the difference between the estimated background and the original input frame, it is successful in detecting moving objects and generates acceptable masks of the foreground. 
\tablresultBMC
\bmcresult
%%%%%%%%%%%%%%%%%%%%%%%%%%%%%%%%%%%%%%%%%%%%%%%%%%%%%%%%%%%%%
\subsection{SBMnet-2016 Dataset}
\label{sec:sbmnet}
SBMnet dataset \cite{jodoin2017extensive} provides a diverse set of 79 videos spanning 8 different categories selected to cover a wide range of detection challenges. These categories consist of basic, intermittent motion, clutter, jitter, illumination changes, background motion, very long with more than 3500 frames, and very short with less than 20 frames. The videos are representative of typical indoor and outdoor visual data captured in surveillance, smart environment.
%, and video database scenarios which makes the level of noise and compression artifacts varies from one video to another. 
Spatial resolutions of the videos vary from $240 \times 240$ to $800\times600$ and their length varies from $6$ to $9370$ frames. Following metrics are utilized to measure the performance. 
\begin{itemize}
    \item Average gray-level error (AGE), average of the gray-level absolute difference between grand truth and the estimated background image.
    \item Percentage of error pixels (pEPs), percentage of number of pixels in estimated background whose value differs from the corresponding pixel in grand truth by more than a threshold with respect to the total number of pixels.
    \item Percentage of clustered error pixels (pCEPS), percentage of error pixels whose 4-connected neighbours are also error pixels with respect to the total number of pixels.
    \item Multi-scale structural similarity index (MSSSIM), estimation of the perceived visual distortion performing at multiple scales.
    \item Peak signal-to-noise-ratio(PSNR), measuring the image quality defined as $10\log_{10}{(\nicefrac{(L-1)^2}{MSE})}$ where $L$ is the maximum grey level value $255$ and $MSE$ is the mean squared error between GT and estimated background.
    \item Color image quality measure (CQM), measuring perceptual image quality defined based on the PSNR values in the single YUV bands through $\text{PSNR}_Y\times R_w+0.5C_w(\text{PSNR}_U+\text{PSNR}_V)$ where PSNR values are in db, $R_w$ and $C_w$ are two coefficients set to $0.9449$ and $0.0551$ respectively.
\end{itemize}
The objective of every background model is an accurate estimate of the background which is equivalent to minimizing AGE, pEPs, pCEP with high perceptual quality equivalent to maximizing the PSNR, MSSSIM, and CQM.
\subsubsection{Performance evaluation and comparison}
\tablresultSBMoverall
\tablresultSBMperCat
\sbmCompareResult
\sbmGlbmGoodVisual
\sbmGlbmBadVisual
For comparison, we analyzed the results obtained by the best performing methods reported on the SBMnet-2016 dataset \footnote{\url{http://scenebackgroundmodeling.net/}} as well as DNN based models compared with our G-LBM model in both quantitative and qualitative manner. \tblref{SBMoverallresults} compares overall performance of our proposed G-LBM model against state-of-the-art background modeling methods including FC-FlowNet \cite{halfaoui2016cnn}, BEWiS \cite{de2017background}, SC-SOBS-C4 \cite{maddalena2016extracting}, BE-AAPSA \cite{ramirez2017temporal}, MSCL \cite{Javed2017}, SPMD \cite{xu2019robust}, LaBGen-OF\cite{laugraud2017memoryless}, FSBE\cite{djerida2019robust}, NExBI\cite{mseddi2019real} with respect to aforementioned metrics.  
Among deep learning approaches in modeling the background practically, other than G-LBM only FC-FlowNet was fully evaluated on SBMnet-2016 dataset. However, the rank of FC-FlowNet is only 20 (see \tblref{SBMoverallresults}) compared with G-LBM and is also outperformed by conventional neural networks approaches like BEWiS, SC-SOBS-C4, and BE-AAPSA. As quantified in \tblref{SBMperCatResult}, G-LBM can effectively capture the stochasticity of the background dynamics in its probabilistic model, which outperforms other methods of background modeling in relative challenges of illumination changes, background motion, and jitter. The qualitative comparison of the G-LBM performance with the other methods is shown in \figref{sbmCompare} for two samples of jitter and background motion categories. However, in clutter and intermittent motion categories that background is heavily filled with clutter or foreground objects are steady for a long time, G-LBM fails in estimating an accurate model compared to other methods that are scene-specific and have special designs for these challenges.
\figref{sbmGlbmGood} visualizes the qualitative results of G-LBM for different challenges in SBMnet-2016 in which it has comparable or superior performance. Cases that G-LBM fails in providing a robust model for the background are also shown in \figref{sbmGlbmBad}, which happen mainly in videos with intermittent motion and heavy clutter.

\section{Conclusion}
    \label{sec:conclusion}
Here, we presented our scene non-specific generative low-dimensional background model (G-LBM) using the framework of VAE for modeling the background in videos recorded by stationary cameras. We evaluated the performance of our model in task of background estimation, and showed how well it adapts to the changes of the background on two datasets of BMC2012 and SBMnet-2016. According to the quantitative and qualitative results, G-LBM outperformed other state-of-the-art models specially in categories that stochasticity of the background is the major challenge such as jitter, background motion, and illumination changes.

%====================================

%\clearpage
\bibliographystyle{splncs04}
\bibliography{egbib}
\end{document}